\documentclass[lettersize,journal]{IEEEtran}
\usepackage{amsmath,amsfonts}
\usepackage{algorithmic}
\usepackage{algorithm}
\usepackage{array}
\usepackage[caption=false,font=normalsize,labelfont=sf,textfont=sf]{subfig}
\usepackage{textcomp}
\usepackage{stfloats}
\usepackage{url}
\usepackage{verbatim}
\usepackage{graphicx}
\usepackage{cite}
\usepackage{xcolor}
\usepackage{multirow}
\usepackage[colorlinks=true]{hyperref}
\begin{document}

\title{LiDAR-based End-to-end Temporal Perception for Vehicle-Infrastructure Cooperation}

\author{Zhenwei Yang$^{1,2,\dag}$,Jilei Mao$^{\dag}$,Wenxian Yang$^{2}$,Yibo Ai$^{1}$,Yu Kong$^{4}$,Haibao Yu$^{2,3,*}$,Weidong Zhang$^{1,*}$ \\
\thanks{$^{\dag}$These authors contributed equally to this work.}
\thanks{$^{*}$is the corresponding author. Contact: wdzhang@ustb.edu.cn; zwdpaper@163.com; yuhaibao94@gmail.com}
\thanks{$^{1}$National Center for Materials Service Safety (NCMS), University of Science and Technology Beijing, Beijing 100083, China.}
\thanks{$^{2}$Institute for AI Industry Research (AIR), Tsinghua University, Beijing 100083, China.}
\thanks{$^{3}$The University of Hong Kong, Hong Kong 999077, China.}
\thanks{$^{4}$Chengfang Financial Information Technology Services CO.LTD, Beijing 100031, China}
}

\maketitle

\begin{center}
    \small
    \textit{This work has been accepted for publication in IEEE Internet of Things Journal.} \\
    \textit{DOI: \href{https://doi.org/10.1109/JIOT.2025.3552526}{10.1109/JIOT.2025.3552526}}
\end{center}

\vspace{1cm}

\begin{abstract}
    Temporal perception, defined as the capability to detect and track objects across temporal sequences, serves as a fundamental component in autonomous driving systems. 
    While single-vehicle perception systems encounter limitations, stemming from incomplete perception due to object occlusion and inherent blind spots, cooperative perception systems present their own challenges in terms of sensor calibration precision and positioning accuracy.
    To address these issues, we introduce LET-VIC, a LiDAR-based End-to-End Tracking framework for Vehicle-Infrastructure Cooperation (VIC).
    First, we employ Temporal Self-Attention and VIC Cross-Attention modules to effectively integrate temporal and spatial information from both vehicle and infrastructure perspectives. 
    Then, we develop a novel Calibration Error Compensation (CEC) module to mitigate sensor misalignment issues and facilitate accurate feature alignment. 
    Experiments on the V2X-Seq-SPD dataset demonstrate that LET-VIC significantly outperforms baseline models. Compared to LET-V, LET-VIC achieves \textbf{+15.0\%} improvement in mAP and a \textbf{+17.3\%} improvement in AMOTA. 
    Furthermore, LET-VIC surpasses representative Tracking by Detection models, including V2VNet, FFNet, and PointPillars, with at least a \textbf{+13.7\%} improvement in mAP and a \textbf{+13.1\%} improvement in AMOTA without considering communication delays, showcasing its robust detection and tracking performance.
    The experiments demonstrate that the integration of multi-view perspectives, temporal sequences, or CEC in end-to-end training significantly improves both detection and tracking performance. All code will be open-sourced.
\end{abstract}


\section{Introduction}
Temporal perception, which encompasses both detection and tracking, is crucial for advanced autonomous driving. 
It enables Connected and Autonomous Vehicles (CAVs) to continuously identify and characterize objects while monitoring their dynamic changes over time, providing a detailed understanding of the dynamic environment. 
However, single-vehicle temporal perception systems face significant challenges, particularly in complex urban settings where occlusions and blind spots can limit the field of view. 
These limitations hinder the ability of CAVs to obtain a complete and reliable understanding of their surroundings.

\begin{figure}[t]
	\centering
	\includegraphics[width=0.48\textwidth]{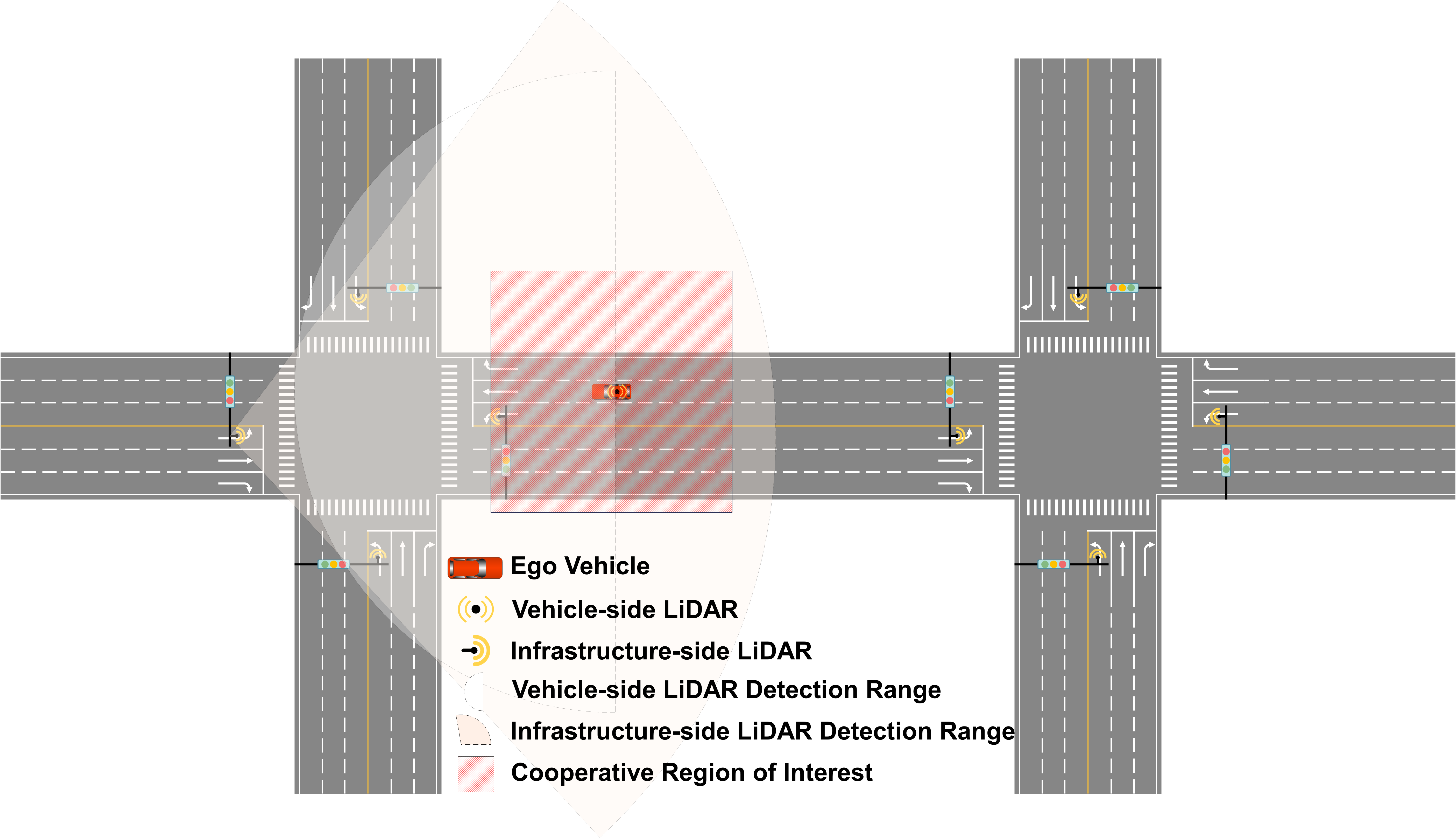}
    \vspace{-10pt}
	\caption{\textbf{Vehicle-Infrastructure Cooperative Diagram.} 
	The red car represents the Ego Vehicle. The vehicle-side LiDAR covers a semicircular area in front of the Ego Vehicle, while the infrastructure-side LiDAR has a fan-shaped coverage area. 
	Through vehicle-infrastructure cooperation, the infrastructure can provide additional perception information to the vehicle.}
	\label{fig:vic-diagram}
\end{figure}

\begin{figure}[t]
	\centering
	\includegraphics[width=0.48\textwidth]{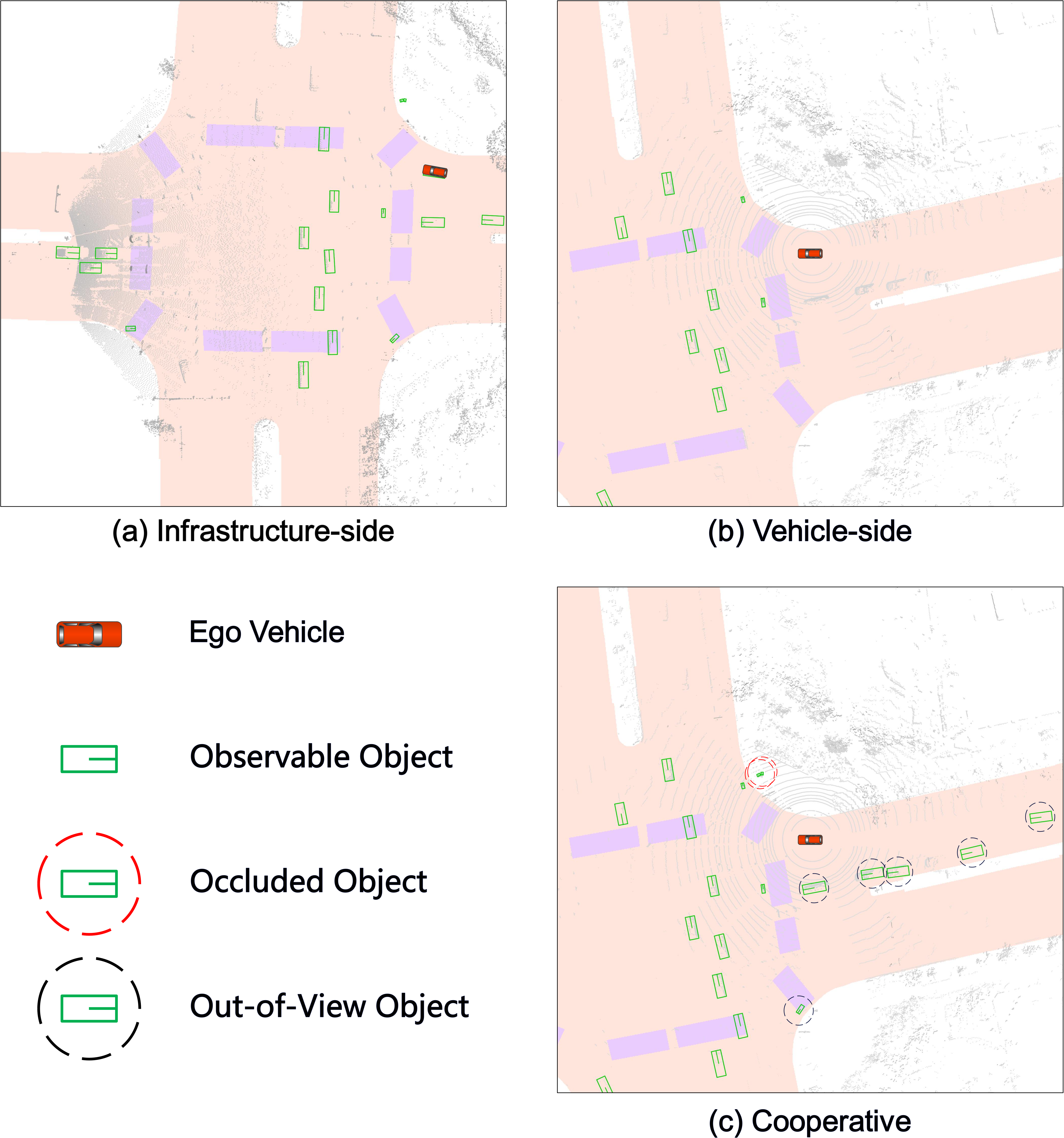}
    \vspace{-10pt}
	\caption{\textbf{Vehicle-Infrastructure Cooperative Perception.} 
	(a) illustrates perception using only infrastructure-side LiDAR, (b) illustrates perception using only vehicle-side LiDAR, and (c) shows the cooperative perception results combining (a) and (b). 
	The red car represents the Ego Vehicle, green boxes indicate detected objects with their orientations, targets within the red dashed circles are those occluded from the vehicle's view, and targets within the black dashed circles are outside the vehicle's perception range.}
	\label{fig:vic-visual}
\end{figure}

Vehicle-Infrastructure Cooperation (VIC), enabled by advancements in Cooperative Intelligent Transportation Systems (C-ITS) and Vehicle-to-Everything (V2X) communication~\cite{chen2017vehicle,gyawali2020challenges,maaloul2021classification}, offers a solution to the limitations of vehicle-only perception. 
Through V2X, data can be exchanged seamlessly between CAVs and infrastructure, allowing the integration of infrastructure-based perception information.
This cooperation enhances situational awareness and fills the gaps caused by occlusions or restricted views on the vehicle side. 
In particular, VIC strengthens temporal perception by providing a more holistic, continuous view of the surroundings, as illustrated in Fig.~\ref{fig:vic-diagram} and Fig.~\ref{fig:vic-visual}.
By combining data from both vehicle and infrastructure sensors, VIC offers more accurate and robust temporal perception compared to standalone vehicle sensors.

Despite the advantages of VIC, current VIC methods are limited in their use of temporal information. 
Existing approaches typically leverage temporal context only during the tracking stage, treating detection as an isolated task for each individual frame. 
This frame-by-frame detection fails to fully exploit the sequential information across frames, which could otherwise improve detection accuracy. 
Moreover, most VIC approaches focus on spatial data fusion, integrating multi-view data from vehicle-side and infrastructure-side sensors at a single moment, without extending this fusion to encompass temporal information. 
Consequently, these approaches are limited by their reliance on instantaneous data, missing out on the benefits of historical context for more informed decision-making.

An end-to-end approach to temporal perception, which utilizes sequential data across frames for both detection and tracking, offers a promising alternative to current VIC methods~\cite{zhang2023motiontrack,meinhardt2022trackformer,zeng2022motr,zhou2020tracking,yin2021center}. 
This paradigm enables a coherent understanding of object dynamics, resulting in a more robust perception system. 
When implemented with LiDAR technology, end-to-end temporal perception becomes even more effective. 
Compared to traditional camera-based systems, LiDAR provides precise distance measurements and maintains reliable performance in adverse weather and low-light conditions, making it an ideal foundation for advanced temporal perception. 
While LiDAR-based end-to-end temporal perception has shown strong potential in vehicle-only systems, extending this approach to a VIC-based framework that fully leverages both spatial and temporal contexts remains underdeveloped. 
Developing such a framework could significantly enhance safety and reliability in autonomous driving applications.

In this study, we propose LET-VIC, the first LiDAR-based End-to-end Tracking framework for VIC, which designed to address major perception challenges in autonomous driving, including limited perception range and occlusion. 
The core innovation lies in simultaneously enhancing perception through two complementary dimensions: spatially fusing infrastructure perspectives with vehicle sensors to expand coverage, and temporally integrating historical observations to resolve occlusion ambiguity, both of which are unified within one efficient framework.
LET-VIC fully utilizes LiDAR point clouds from both vehicle-side and infrastructure-side sensors, incorporating temporal data to greatly enhance object detection and tracking in complex traffic environments. 
The framework features an advanced feature fusion strategy that uses a VIC cross-attention mechanism to seamlessly integrate Bird's Eye View (BEV) features from the infrastructure-side into the vehicle-side BEV representation. 
This dual-fusion approach, which combines cross-sensor spatial fusion with temporal evolution modeling, enables the system to overcome both physical obstructions and single-frame sensing limitations.
Furthermore, We employ a Calibration Error Compensation (CEC) module to compensate for calibration errors between vehicle-side and infrastructure-side sensors, delivering precise and robust 3D object perception. 
Compared to existing approaches, LET-VIC is novel in its ability to unify spatial and temporal data fusion in an end-to-end manner. This ensures not only better detection and tracking performance but also improved robustness to environmental complexity and sensor misalignments, making it a critical step forward for autonomous driving applications.

The contributions are summarized as follows:
\begin{itemize}
    \item We introduce LET-VIC, a LiDAR-based End-to-end Tracking framework for VIC that leverages a novel VIC cross-attention mechanism to fuse BEV features from both vehicle and infrastructure side sensors, providing a unified solution for 3D object perception in complex traffic scenarios.
    \item We propose a self-supervised method to optimize spatial error correction during coordinate transformation. By learning offsets $\Delta x$ and $\Delta y$ within BEV queries, this method corrects misalignments caused by calibration errors, enhancing robustness against coordinate inaccuracies.
    \item LET-VIC consistently outperforms existing cooperative temporal perception methods on the V2X-Seq-SPD dataset~\cite{yu2023v2x}. Our experimental design also included evaluations under various delay conditions ranging from 0 ms to 300 ms, demonstrating LET-VIC's strong stability and adaptability, even though it was not explicitly optimized for delay handling.
\end{itemize}

The remainder of this paper is organized as follows. Section II reviews the related works relevant to this study. In Section III, we present LET-VIC, our proposed LiDAR-based end-to-end framework for VIC tracking.
Section IV details the experimental results and analysis. Finally, Section V concludes the paper.

\section{Related Works}
\subsection{Cooperative Perception} 
Cooperative perception in autonomous driving enhances perception capabilities to address challenges such as occlusion, long-distance perception, and sensor failures that single-vehicle perception may struggle with. 
Recent cooperative perception research primarily focuses on two types: Vehicle-to-Vehicle (V2V) and Vehicle-to-Infrastructure (V2I) cooperation~\cite{sehla2022resource,cheng2022integrated,jang2023study,lu2024integrated}.

In V2V cooperation, different fusion approaches have been explored. Early Fusion techniques, as seen in Cooper~\cite{chen2019cooper} and DiscoNet~\cite{li2021learning}, 
integrate raw sensor data from multiple vehicles, improving detection accuracy and extending perception range. 
Feature Fusion approaches, such as OPV2V~\cite{xu2022opv2v}, F-Cooper~\cite{chen2019f}, V2VNet~\cite{wang2020v2vnet}, and COOPERNAUT~\cite{cui2022coopernaut}, 
combine intermediate-level features from multiple vehicles for enhanced perception. 
Notably, How2comm~\cite{yang2023how2comm} focuses on transmitting BEV features extracted from point clouds to enhance cooperative 3D object detection.

On the other hand, V2I cooperation has been explored through models such as V2X-ViT~\cite{xu2022v2x}, FFNet~\cite{yu2023flow}, CTCE~\cite{zhong2024leveraging}, V2X-Graph~\cite{ruan2023learning} and UniV2X{yu2024end}, 
which adopt a feature fusion approach to integrate data from both vehicles and infrastructure to provide broader environmental awareness. 

These diverse approaches underscore significant efforts to leverage collective sensing capabilities. These capabilities help overcome the inherent limitations of individual vehicle sensors, ultimately enhancing the reliability and safety of autonomous driving systems.
While many studies focus on specific perception tasks, integrating cooperative strategies into broader end-to-end systems remains a promising direction for future research, offering the potential for more comprehensive system performance in complex driving environments.

\subsection{End-to-End Temporal Perception}
The paradigm of End-to-End Temporal Perception has redefined sequential object perception by shifting from traditional tracking-by-detection to a unified model which has been proven to simultaneously enhance both detection and tracking accuracy~\cite{zeng2022motr,zhang2022mutr3d,zhang2023motrv2,yu2023motrv3,lin2023sparse4dv3,li2023end}. 
MotionTrack~\cite{zhang2023motiontrack} introduces a transformer-based end-to-end temporal perception algorithm utilizing LiDAR-camera fusion, setting a new benchmark for tracking multiple object classes with multimodal sensor inputs. 
TrackFormer~\cite{meinhardt2022trackformer} and MOTR~\cite{zeng2022motr} reformulate temporal perception as a set prediction problem, leveraging transformer architectures to maintain track identity and spatiotemporal trajectories without requiring separate motion or appearance modeling.
CenterTrack~\cite{zhou2020tracking} simplifies the process by performing detection and tracking simultaneously, significantly improving temporal perception efficiency and accuracy. 
CenterPoint~\cite{yin2021center}, a LiDAR-based 3D object detection and tracking framework, introduces an anchor-free approach that directly predicts object center points and regresses their attributes.

However, the end-to-end framework has not been thoroughly explored in the LiDAR-based VIC field. HYDRO-3D~\cite{meng2023hydro} improve detection accuracy by leveraging temporal information and collaborative perception, yet they lack a comprehensive evaluation of tracking performance.
MOT-CUP~\cite{su2024collaborative} proposes a joint perception post-processing framework validated under ideal simulation V2X-Sim dataset~\cite{li2022v2x}. 
However, the effectiveness of real-time collaborative perception and tracking is significantly impacted by transmission delays, data timestamp alignment, and the precision of coordinate transformation matrices, all of which are critical and non-negligible factors. 
These issues substantially increase the complexity of hyperparameter tuning for such post-processing algorithms.
In contrast, end-to-end training addresses these challenges by employing sophisticated network designs, eliminating the need for manual hyperparameter configuration while achieving greater robustness.
\cite{su2024cooperative} and \cite{chiu2024probabilistic} introduce novel 3D multi-object cooperative tracking algorithms based on a post-fusion framework, evaluated on the real-world V2V4Real dataset~\cite{xu2023v2v4real}. 
However, these approaches solely focus on optimizing the tracking component and fail to improve detection accuracy as same as MOT-CUP~\cite{su2024collaborative}.

To bridge these gaps, we introduce a novel LiDAR-based end-to-end temporal perception framework for VIC, which leverages multi-view fusion between vehicle-side and infrastructure-side sensors. 
By integrating complementary spatial and temporal information, this framework significantly enhances perception accuracy and robustness, marking a significant advance in cooperative temporal perception.

\subsection{BEV-based Perception}
BEV-based perception has emerged as a powerful method for enhancing object detection and tracking in autonomous driving by enabling accurate, unified spatial representations from multi-view and multimodal data. 
LSS~\cite{philion2020lift} pioneered the use of depth estimation to project image features into BEV space, effectively addressing the challenge of fusing multi-view image data. 
BEVDet4d~\cite{huang2022bevdet4d} further optimized this process, while BEVDepth~\cite{li2023bevdepth} and BEVStereo~\cite{li2023bevstereo} improved the precision of image-to-BEV feature projection. 
Unlike the BEVDet series of methods, BEVFormer~\cite{li2022bevformer} introduced spatial cross-attention to fuse features from different camera views in BEV space.
Methods like Simple-BEV~\cite{harley2023simple}, Fast-BEV~\cite{li2024fast}, and MatrixVT~\cite{zhou2023matrixvt} have made BEV feature generation more efficient. 
Additionally, recent works such as Is-Fusion~\cite{yin2024fusion}, BEVFusion~\cite{liu2023bevfusion}, and FusionFormer~\cite{hu2023fusionformer} have focused on multi-modal data integration, combining image and LiDAR data for more accurate perception.
These advancements laid the foundation for more sophisticated end-to-end autonomous driving systems~\cite{li2023uniformer,hu2023planning,zhang2024graphad,ye2023fusionad,sun2024sparsedrive,yu2024end}. 
We adopt BEV-based perception as it allows our system to integrate multi-sensor inputs seamlessly and accurately, supporting more reliable and robust performance in dynamic driving environments.

\begin{figure*}[ht]
	\centering
	\includegraphics[width=1.0\textwidth]{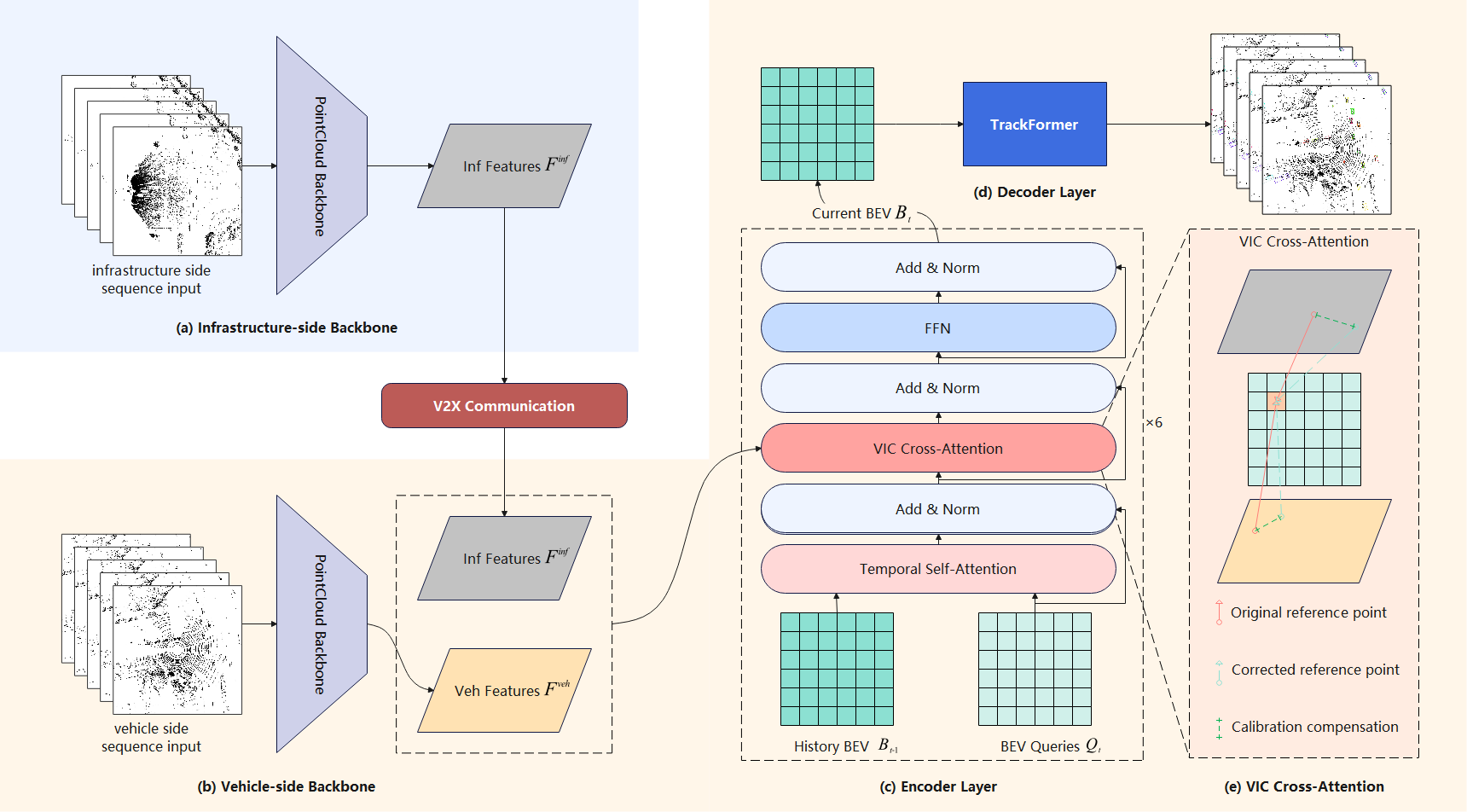}
  \vspace{-10pt}
	\caption{\textbf{Architecture of LET-VIC.} The diagram illustrates the steps involved in the LET-VIC framework. (a) and (b) Both the infrastructure-side and vehicle-side employ PointPillars~\cite{lang2019pointpillars} to extract LiDAR point cloud features. Then the infrastructure-side features are transmitted to the vehicle side through V2X communication and integrated into the VIC cross-attention module along with the vehicle-side features. (c) The BEV encoder of LET-VIC is inspired by BEVFormer. (d) The decoder layers of LET-VIC employ TrackFormer. (e) The VIC Cross-Attention module fuses the Bird's Eye View (BEV) features from both the infrastructure and vehicle sides, compensating for calibration errors.}
	\label{fig:LET-VIC-framework}
\end{figure*}

\section{Method}
This section describes the LET-VIC method utilized in our proposed LiDAR-based end-to-end tracking framework for VIC.
LET-VIC's key novelty lies in dynamically combining LiDAR data from both vehicle and infrastructure perspectives through a novel VIC Cross-Attention mechanism. 
This design enables more accurate object detection and tracking by addressing challenges such as sensor calibration errors and limited individual sensor coverage. 
In essence, LET-VIC enhances environmental awareness, enabling safer and more efficient autonomous driving.
The end-to-end design further simplifies the traditional tracking pipeline, combining object detection and tracking into a seamless, jointly optimized framework, improving efficiency and accuracy.
As illustrated in Fig.~\ref{fig:LET-VIC-framework}, LET-VIC leverages PointPillars~\cite{lang2019pointpillars} for feature extraction from both vehicle and infrastructure LiDAR data. 
The features are then fused using the VIC Cross-Attention module to enhance feature integration on the vehicle side, 
followed by detection and tracking using the MOTR approach~\cite{zeng2022motr}.
Additionally, the framework addresses the significant impact of sensor calibration errors on detection and tracking accuracy through an online sensor calibration compensation module.
We begin by defining the VIC temporal perception problem and providing relevant background information. 
Following this, we present our approach to generating BEV features and detail the methodology for effectively fusing multi-view data. 
We then address the significant challenge of sensor calibration errors and propose a compensation mechanism to mitigate their impact. 
Subsequently, we present our end-to-end temporal perception strategy. 
Finally, we discuss the evaluation metrics employed to assess the overall effectiveness of LET-VIC.

\subsection{Task Description}\label{sec:task}
VIC Temporal Perception encompasses the dual objectives of detecting and tracking objects using sequential point cloud data collected from both vehicle and infrastructure sensors. 
This collaborative approach enhances the overall perception of the environment, allowing for more accurate identification and monitoring of objects within the vicinity of the ego vehicle~\cite{yu2023v2x}. 
The study focuses specifically on point cloud data, which contains critical depth information obtained from LiDAR sensors, facilitating a robust understanding of the spatial and temporal characteristics of the objects.

In the context of VIC Detection, the system is designed to identify various objects within the environment. 
The detection process relies on sequential point cloud data captured over time, allowing for the recognition of objects based on their spatial features. 
This includes determining the type of objects present, their spatial locations, dimensions and orientations. 
The integration of data from both vehicle and infrastructure sensors is crucial, as it provides a comprehensive view of the surroundings, 
improving detection accuracy and robustness against occlusions or limitations inherent to individual sensors.

For the tracking aspect, VIC Tracking focuses on the continuous monitoring of detected objects as they move through the environment. 
This involves maintaining a unique tracking ID for each object, which allows the system to consistently identify and follow these objects over time. 
The tracking process is particularly challenging due to factors such as sensor noise, changes in object appearance, and the dynamic nature of the environment. 
By leveraging sequential data from both the vehicle and infrastructure sides, the system can achieve more stable and reliable tracking of moving objects, even in complex scenarios.

The input of LET-VIC consists of two components:
\begin{itemize}
    \item Vehicle sequential point clouds ${PC_v(t_v^{'})|t_v^{'}\leq t_v}$ and their calibration matrix ${M_v(t_v^{'})|t_v^{'}\leq t_v}$, captured at and prior to time $t_v$. Here, $PC_v(\cdot)$ denotes the vehicle-side sensor capturing function.
    \item Infrastructure sequential point clouds ${PC_i(t_i^{'})|t_i^{'}\leq t_i}$ and their calibration matrix ${M_i(t_i^{'})|t_i^{'}\leq t_i}$, captured before or at time $t_i$. Here, $PC_i(\cdot)$ denotes the infrastructure-side sensor capturing function. Note that $t_i$ is earlier than $t_v$ ($t_i < t_v$) due to communication delays.
\end{itemize}

The outputs of VIC Temporal Perception include the category, location, dimension, orientation, and unique tracking ID of each object in the region of interest surrounding the ego vehicle at time $t_v$. 
The corresponding ground truth consists of the set of tracked objects detected and tracked by any of the cooperative-view sensors at time $t_v$, which can be formulated as:
\begin{equation}
\setlength\abovedisplayskip{0.05cm}
\setlength\belowdisplayskip{0.15cm}
GT = (GT_{v} \cup GT_{i}) \cap ROI_{veh},
\end{equation}
where $GT_{v}$ is the ground truth from the vehicle side sensors, $GT_{i}$ is the ground truth from the infrastructure side sensors, and $ROI_{veh}$ is the ego-vehicle region of interest.

\begin{figure}[t]
	\centering
	\includegraphics[width=0.24\textwidth]{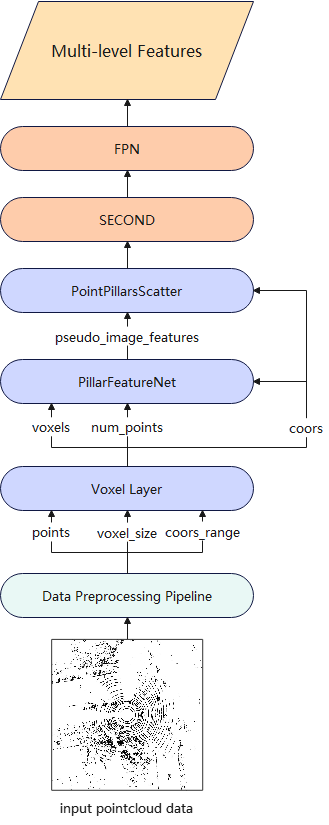}
  \vspace{-10pt}
	\caption{\textbf{PointCloud Backbone.} The PointCloud Backbone extract Bird's Eye View (BEV) features from point cloud data. The pipeline starts with data preprocessing, voxelization, and feature extraction via PillarFeatureNet. Pseudo-image features are then processed by SECOND and FPN, producing multi-level features for object perception at various scales.}
	\label{fig:pc backbone}
  \vspace{-10pt}
\end{figure}

\subsection{BEV Feature Generation}
Figure~\ref{fig:pc backbone} provides a detailed illustration of the Pointcloud Backbone used in both the vehicle-side and infrastructure-side components of the LET-VIC framework, corresponding to parts (a) and (b) in Fig.~\ref{fig:LET-VIC-framework}. 
The diagram shows how the PointPillars framework~\cite{lang2019pointpillars} is employed to process LiDAR point clouds on both sides.
The raw point cloud data is first organized into pillars—grids that aggregate spatial information—enabling efficient feature extraction for subsequent processing.

The pipeline for BEV feature generation begins by preprocessing the 3D point clouds (x, y, z). 
In the voxelization step, the point clouds are discretized into 3D grid cells (voxels), with distinct regions of interest (ROI) for vehicle and infrastructure sides. 
The vehicle side ROI spans x: [-51.2, 51.2] meters, y: [-51.2, 51.2] meters, and z: [-5.0, 3.0] meters, while the infrastructure side ROI covers x: [0, 102.4] meters, y: [-51.2, 51.2] meters, and z: [-5.0, 3.0] meters.

Voxelization uses a uniform voxel size of 0.2 meters along the x and y axes, and 8 meters along the z-axis. 
As a result, the vehicle-side grid is $512\times512$, and the infrastructure-side grid is also $512\times512$. Each voxel stores its corresponding points and their coordinates.

Following voxelization, PillarFeatureNet extracts high-dimensional features from the voxels, transforming the point cloud into a 2D BEV representation. 
These pseudo-image features are scattered using the PointPillarsScatter layer and are further refined by SECOND convolutional layers. 
Finally, a Feature Pyramid Network (FPN) integrates multi-scale features, outputting a set of multi-level features to enhance object perception capabilities across different spatial resolutions.

\subsection{BEV Feature Transmission}
The block diagram between parts (a) and (b) of Fig.~\ref{fig:LET-VIC-framework} illustrates the transmission of BEV features from the infrastructure side to the vehicle side.
In V2X communication, three primary data types are transmitted: raw data, instance-level perception data, and feature-level intermediate data. Each type incurs different transmission costs and offers varying levels of information retention.

\begin{itemize}
    \item Raw data includes unprocessed sensor information, like LiDAR point clouds or images. It has the highest transmission cost due to its large volume but retains the most information as it is unfiltered and uncompressed.
    \item Instance-level perception data consists of processed outputs like detected objects and attributes or queries that contain perception results but are not yet decoded~\cite{fan2023quest}. It has the lowest transmission cost due to high compression, but this also leads to the highest information loss because it transmits only selected information.
    \item Feature-level intermediate data comprises key features extracted from raw data, such as BEV features~\cite{chen2023transiff}. It balances transmission costs and information retention.
\end{itemize}

Our V2X communication framework prioritizes transmitting BEV features due to the trade-off between transmission cost and information retention. BEV features capture essential spatial relationships and environmental context with reduced data volume, ensuring critical information for accurate perception and decision-making is retained. This approach significantly reduces the required communication bandwidth compared to raw data transmission. Additionally, the VIC Cross-Attention mechanism effectively integrates BEV features from both vehicle and infrastructure sides.

\subsection{BEV Feature Fusion}
The deformable attention mechanism plays a critical role in LET-VIC's ability to dynamically fuse features from multiple viewpoints. 
Traditional attention mechanisms can struggle with multi-view fusion due to high computational costs and difficulties in aligning spatial features across different perspectives. 
In contrast, deformable attention significantly reduces computational overhead by focusing on a sparse set of key features, known as reference points, 
and dynamically adjusting its focus based on local feature distributions.

In LET-VIC, the VIC Cross-Attention Module employs deformable attention to integrate BEV features from both vehicle-side and infrastructure-side sensors. 
The process begins by projecting BEV queries to each perspective and generating initial reference points. 
A learnable network then predicts calibration adjustments to refine these reference points, compensating for any misalignments between the two perspectives. 
Deformable attention is subsequently applied around the refined points, enabling the model to sample key features efficiently. 
This approach ensures that only the most relevant features from both viewpoints are fused, enhancing the efficiency and accuracy of feature integration while accounting for sensor misalignments.

\begin{figure}[t]
	\centering
	\includegraphics[width=0.48\textwidth]{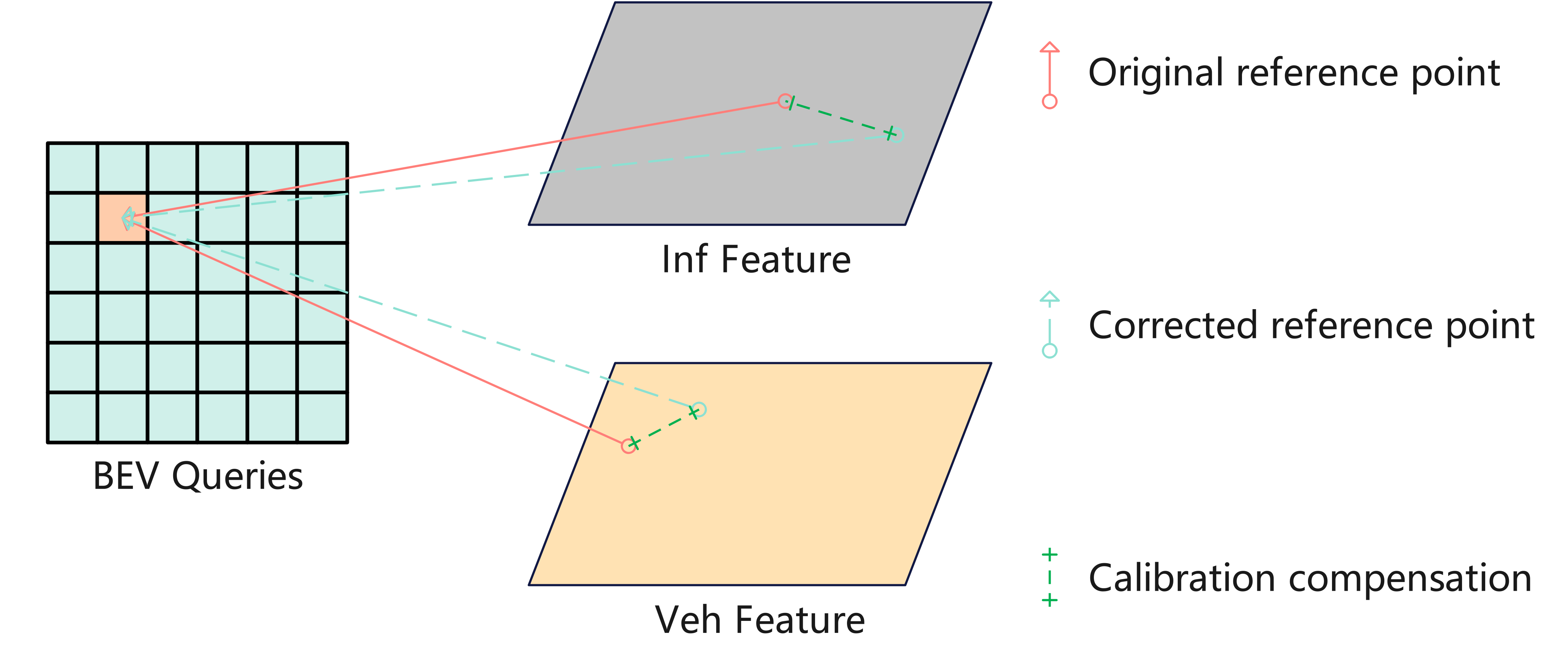}
  \vspace{-10pt}
	\caption{\textbf{VIC Cross-Attention.} The VIC Cross-Attention module fuses the Bird's Eye View (BEV) features from both the infrastructure and vehicle sides, compensating for calibration errors. First, We project BEV queries to both the infrastructure side and vehicle side features to get the original reference point. Then, we use a learnable network to predict the calibration compensation to refine the original reference point. Finally, we use deformable attention sampling around the corrected reference point to generate the respective BEV feature.}
	\label{fig:vic cross-attention}
  \vspace{-10pt}
\end{figure}

The ability to adaptively focus on critical features enables the deformable attention mechanism to excel in complex, multi-view tracking tasks where spatial alignment may vary due to sensor calibration errors.
Figure~\ref{fig:vic cross-attention} corresponds to part (e) in Fig.~\ref{fig:LET-VIC-framework} and illustrates the VIC Cross-Attention mechanism, showcasing its ability to dynamically integrate data across different sensors. 
The process of VIC Cross-Attention (VICCrossAttn) can be formulated as follows:
\begin{align}\label{eq:vic cross attention}
\setlength\abovedisplayskip{0.05cm}
\setlength\belowdisplayskip{0.15cm}
&\text{VICCrossAttn}(Q^{bev},F^{veh},F^{inf},P^{veh}, P^{inf})=\notag\\
&\sum_{i=1}^{N_{ref}}\left[\frac{1}{1+\mathcal{M}_{i}^{inf}}\left(\text{MSDeformAttn}(Q_i^{bev},P_i^{veh},F^{veh})\right.\right.\notag\\
&+\left.\left.\mathcal{M}_{i}^{inf}\cdot\text{MSDeformAttn}(Q_i^{bev},P_i^{inf},F^{inf})\right)\right],
\end{align}
where $i$ indexes the reference points, and $N_{ref}$ is the total reference points
for BEV Query $Q^{bev}$. ${P}_i^{veh}\in[0,1]^2$ are the normalized coordinates of the $i$-th reference points on the vehicle side. ${P}_i^{inf}\in[0,1]^2$ are the normalized coordinates of the $i$-th reference points on the infrastructure side. $\mathcal{M}^{inf}$ is the mask matrix ensuring that any reference point projected onto the infrastructure side multi-level LiDAR BEV feature is excluded if it falls outside the feature's bounds.

The multi-scale deformable attention (MSDeformAttn) function is defined as:
\begin{align}\label{eq:msdeformattn}
\setlength\abovedisplayskip{0.05cm}
\setlength\belowdisplayskip{0.15cm}
&\text{MSDeformAttn}(Q_i,{P}_i,\{F^l\}_{l=1}^L)\notag\\
&=\sum_{m=1}^M{W_m}\Bigg[\sum_{l=1}^L\sum_{k=1}^K A_{mlqk}\cdot{W'_m}F_i^l(\mathcal{R}_{l}({P}_i)+\Delta{p_{mlqk}})\Bigg],
\end{align}
Where $i\in\Omega_i$ indexes a query element with representation feature $Q_i\in\mathbb{R}^C$, where $C$ is the feature dimension. $m$ indexes the attention head, $l$ indexes the input feature level, and $k$ indexes the sampling point. $\Delta{p_{mlqk}}$ represents the sampling offset, and $A_{mlqk}$ denotes the attention weight of the $k$-th sampling point in the $l$-th feature level and the $m$-th attention head. The scalar attention weight $A_{mlqk}$ is normalized by $\sum_{l=1}^L\sum_{k=1}^K{A_{mlqk}}=1$. ${P}_i\in[0,1]^2$ are the normalized coordinates for clarity in scale formulation. The function $\mathcal{R}_l({P}_i)$ re-scales the normalized coordinates ${P}_i$ to the input feature map of the $l$-th level. $\Delta p{mlqk}\in\mathbb{R}^2$ are 2-dimensional real numbers with an unconstrained range. $W_m'\in\mathbb{R}^{C/M \times C}$ and $W_m\in\mathbb{R}^{C\times{C/M}}$ are learnable weights for the $m$-th attention head.

\subsection{Calibration Error Compensation}
Calibration errors in sensors, arising from precision variations, environmental factors, and calibration method uncertainties, pose significant challenges to perception accuracy in autonomous driving systems. 
Compensating for sensor calibration errors is crucial for LET-VIC as it ensures accurate data fusion and reliable cooperative perception performance.
Therefore, we implemented CEC module to optimize spatial error correction during coordinate transformation, delivering precise and robust temporal perception.
The CEC module is integrated into the VIC Cross-Attention mechanism, which corresponds to part (e) in Fig.~\ref{fig:LET-VIC-framework}.

Initially, the positions of vehicle-side or infrastructure-side features corresponding to each point in the BEV features are determined through coordinate system transformation. 
Subsequently, CEC employs a set of learnable parameters that adapt during training to learn adaptive offsets, which include ${\Delta x}^{inf}$, ${\Delta y}^{inf}$, ${\Delta x}^{veh}$, and ${\Delta y}^{veh}$. 
Finally, the positions of vehicle-side or roadside features corresponding to each BEV feature are adjusted by adding this set of offsets, forming compensated reference points as shown in (\ref{eq:veh error offsets}) and (\ref{eq:inf error offsets}). 
Then, we normalize the original reference point coordinates according to the specified point cloud range, as detailed in (\ref{eq:normalized veh error offsets}) and (\ref{eq:normalized inf error offsets}).
\begin{align}\label{eq:veh error offsets}
\setlength\abovedisplayskip{0.05cm}
\setlength\belowdisplayskip{0.15cm}
\begin{bmatrix}x_i^{veh}\\y_i^{veh}\end{bmatrix}=R_{bev2veh}\cdot{Ref}_i^{bev}+T_{bev2veh}+\begin{bmatrix} {\Delta x}_i^{veh} \\ {\Delta y}_i^{veh} \end{bmatrix},
\end{align}
\begin{align}\label{eq:inf error offsets}
\setlength\abovedisplayskip{0.05cm}
\setlength\belowdisplayskip{0.15cm}
\begin{bmatrix}x_i^{inf}\\y_i^{inf}\end{bmatrix}=R_{bev2inf}\cdot{Ref}_i^{bev}+T_{bev2inf}+\begin{bmatrix} {\Delta x}_i^{inf} \\ {\Delta y}_i^{inf} \end{bmatrix},
\end{align}
\begin{align}\label{eq:normalized veh error offsets}
\setlength\abovedisplayskip{0.05cm}
\setlength\belowdisplayskip{0.15cm}
{P}_i^{veh}=\begin{bmatrix} \cfrac{x_i^{veh}-x_{min}^{veh}}{x_{max}^{veh}-x_{min}^{veh}} & \cfrac{y_i^{veh}-y_{min}^{veh}}{y_{max}^{veh}-y_{min}^{veh}} \end{bmatrix},
\end{align}
\begin{align}\label{eq:normalized inf error offsets}
\setlength\abovedisplayskip{0.05cm}
\setlength\belowdisplayskip{0.15cm}
{P}_i^{inf}=\begin{bmatrix} \cfrac{x_i^{inf}-x_{min}^{inf}}{x_{max}^{inf}-x_{min}^{inf}} & \cfrac{y_i^{inf}-y_{min}^{inf}}{y_{max}^{inf}-y_{min}^{inf}} \end{bmatrix},
\end{align}
Where $x_i^{veh}$, $y_i^{veh}$ represents the $i$-th original coordinates of the reference point on the vehicle-side and $x_i^{inf}$, $y_i^{inf}$ represents the $i$-th original coordinates of the reference point on the infrastructure-side. $R_{bev2inf}$ represents the rotation matrix from the vehicle side LiDAR coordinate system to the pseudo-image coordinate system of infrastructure-side point cloud BEV features, and $T_{bev2inf}$ represents the translation matrix from the vehicle side LiDAR coordinate system to the pseudo-image coordinate system of infrastructure-side point cloud BEV features. ${\Delta x}_i^{inf}$ and ${\Delta y}_i^{inf}$ represent the calibration errors corresponding to the infrastructure side at $i$-th reference point; ${Ref}_i^{veh}$ represents the $i$-th reference points on the pseudo-image of vehicle side point cloud BEV features, $R_{bev2veh}$ represents the rotation matrix from the vehicle side LiDAR coordinate system to the pseudo-image coordinate system of vehicle side point cloud BEV features, and $T_{bev2veh}$ represents the translation matrix from the vehicle side LiDAR coordinate system to the pseudo-image coordinate system of vehicle side point cloud BEV features; ${\Delta x}_i^{veh}$ and ${\Delta y}_i^{veh}$ represent the calibration errors corresponding to the vehicle side at $i$-th reference point. $x_{min}^veh$, $x_{max}^veh$ are the minimum and maximum values of the vehicle side point cloud range in the x-direction, $y_{min}^veh$, $y_{max}^veh$ are the minimum and maximum values of the vehicle side point cloud range in the y-direction. $x_{min}^inf$, $x_{max}^inf$ are the minimum and maximum values of the infrastructure side point cloud range in the x-direction, $y_{min}^inf$, $y_{max}^inf$ are the minimum and maximum values of the infrastructure side point cloud range in the y-direction. ${P}_i^{veh}$, ${P}_i^{inf}$ are the input parameters defined in (\ref{eq:vic cross attention}).

\subsection{End-to-End Tracking}
The end-to-end tracking capability of LET-VIC, shown in part (d) of Fig.~\ref{fig:LET-VIC-framework}, is powered by an advanced architecture inspired by the TrackFormer framework~\cite{zeng2022motr,meinhardt2022trackformer,hu2023planning}, which unifies object detection and tracking tasks within a single Transformer-based pipeline. 
This integration bypasses the conventional two-step tracking-by-detection approach, where detection is followed by post-processing for tracking. 
Instead, LET-VIC directly associates objects across frames during detection, allowing for joint optimization of both tasks.

Within LET-VIC, detect queries and track queries play distinct but complementary roles in the TrackFormer decoder. 
Detect queries are responsible for identifying new objects in each frame by interacting with the current frame's features, focusing on objects not previously tracked. 
Track queries, on the other hand, are used to maintain object identities across frames, carrying information from previous frames to ensure temporal consistency by associating the same object across consecutive frames. 
This combination allows for simultaneous detection and tracking, ensuring seamless object tracking throughout sequences.

LET-VIC's end-to-end design benefits from a joint loss function that simultaneously optimizes detection and tracking. 
The loss function includes components for object classification, bounding box regression, and identity preservation. 
This comprehensive approach enhances overall accuracy and ensures that objects are correctly identified and tracked even in complex and dynamic environments.

\subsection{Evaluation Metrics and Analysis}\label{sec:metrics}
To evaluate LET-VIC's performance, we selected a range of metrics that comprehensively assess both detection and tracking accuracy~\cite{caesar2020nuscenes,geiger2013vision,weng20203d}. 
These metrics are standard in 3D tracking benchmarks and provide a balanced view of system performance under various conditions:

\begin{itemize}
  \item \textbf{mAP (Mean Average Precision):} mAP is a key metric for evaluating object detection performance. 
  It measures how accurately detected objects match ground-truth objects across different recall levels, typically using distance thresholds. 
  For each distance threshold, precision-recall curves are generated, and the area under each curve gives the Average Precision (AP). 
  The final mAP score is the mean of AP values across all classes and distance thresholds. 
  A higher mAP indicates better detection quality. The formulations for AP and mAP are:
  \begin{align}\label{eq:map}
    \setlength\abovedisplayskip{0.05cm}
    \setlength\belowdisplayskip{0.15cm}
    \text{AP}=\int_{0}^{1}\text{precision}(r)dr,
  \end{align}
  \begin{align}\label{eq:map}
    \setlength\abovedisplayskip{0.05cm}
    \setlength\belowdisplayskip{0.15cm}
    \text{mAP}=\cfrac{1}{N}\sum_{c=1}^{N}\text{AP}_c,
  \end{align}
  Where $r$ represents recall levels from 0 to 1, $N$ is the number of classes, and $\text{AP}_c$ is the average precision for class $c$. In practice, the area under the PR curve is often approximated by taking the precision at discrete recall levels.
  \item \textbf{AMOTA (Average Multi-Object Tracking Accuracy):} AMOTA is a central metric for evaluating tracking accuracy in multi-object scenarios. 
  It builds on Multi-Object Tracking Accuracy (MOTA), which assesses tracking accuracy by accounting for false positives, false negatives, and identity switches between tracked objects and ground truth.
  AMOTA averages MOTA scores across different difficulty levels and recall thresholds, providing an overall measure of tracking accuraciy.
  Higher AMOTA values are preferred, as they reflect consistent tracking accuracy across recall levels. The formulations for MOTA and AMOTA are:
  \begin{align}\label{eq:mota}
    \setlength\abovedisplayskip{0.05cm}
    \setlength\belowdisplayskip{0.15cm}
    \text{MOTA}=1-\cfrac{\sum_{t}(\text{FP}_t+\text{FN}_t+\text{IDS}_t)}{\sum_{t}(\text{GT}_t)},
  \end{align}
  \begin{align}\label{eq:amota}
    \setlength\abovedisplayskip{0.05cm}
    \setlength\belowdisplayskip{0.15cm}
    \text{AMOTA}=\cfrac{1}{R}\sum_{r=1}^{R}\text{MOTA}_r,
  \end{align}
  Where $\text{FP}_t$ is the number of false positives at time $t$, $\text{FN}_t$ is the number of false negatives at time $t$, $\text{IDS}_t$ is the number of ID switches at time $t$, $\text{GT}_t$ is the total number of groundtruth objects at time $t$, 
  $R$ is the number of recall levels, and $\text{MOTA}_r$ is the MOTA score at recall level $r$.
  \item \textbf{AMOTP (Average Multi-Object Tracking Precision):} AMOTP is an essential metric for evaluating tracking precision in multi-object scenarios.
  It builds on Multi-Object Tracking Precision (MOTP), which measures the spatial alignment between tracked and ground-truth object positions by calculating the average positional error.
  AMOTP averages MOTP scores across different difficulty levels and recall thresholds, providing an overall measure of tracking precision.
  Lower AMOTP values are preferred, as they reflect consistent tracking precision across recall levels. The formulations for MOTP and AMOTP are:
  \begin{align}\label{eq:motp}
    \setlength\abovedisplayskip{0.05cm}
    \setlength\belowdisplayskip{0.15cm}
    \text{MOTP}=\cfrac{\sum_{i,t}d_{i,t}}{\sum_{t}(c_t)},
  \end{align}
  \begin{align}\label{eq:amotp}
    \setlength\abovedisplayskip{0.05cm}
    \setlength\belowdisplayskip{0.15cm}
    \text{AMOTP}=\cfrac{1}{R}\sum_{r=1}^{R}\text{MOTP}_r,
  \end{align}
  Where $d_{i,t}$ is the positional error for the $i$-th matched object at time $t$, $c_t$ is the number of successfully matched object pairs at time $t$, 
  $R$ is the number of recall levels, and $\text{MOTP}_r$ is the MOTP score at recall level $r$.
\end{itemize}

These metrics offer a thorough evaluation of LET-VIC's detection and tracking performance under varied conditions. 
We select mAP as our primary detection metric because it effectively measures the accuracy of object identification, providing insight into the model's ability to detect objects with precision. 
For tracking, we emphasize AMOTA and AMOTP as they capture tracking accuracy and spatial precision across multiple recall thresholds.
These metrics provide a comprehensive assessment of LET-VIC's robustness and consistency, especially in complex, real-world environments where accurate object identification and stable tracking are critical.

\section{Experiments}
In this section, we evaluate the performance of LET-VIC on V2X-Seq-SPD dataset~\cite{yu2023v2x}. 
For better comparison, we also develop and reproduce four methods, including one single-view end-to-end method: LET-V, 
and three multi-view tracking-by-detection methods: 
V2VNet~\cite{wang2020v2vnet}, FFNet~\cite{yu2023flow}, and PointPillars~\cite{lang2019pointpillars} with tracking method AB3DMOT~\cite{weng20203d}. 
The experimental results indicate that our method can significantly improve performance over the compared methods. 

\subsection{Experiment Settings}
\subsubsection{Dataset}\label{sec:let-vic datasets}
We conduct experiments on two datasets: the real-world \textbf{V2X-Seq-SPD} dataset~\cite{yu2023v2x} and the simulated \textbf{V2X-Sim} dataset~\cite{li2022v2x}, collected from Carla.  

\textbf{V2X-Seq-SPD}~\cite{yu2023v2x} is a real-world dataset designed for VIC temporal perception. 
It consists of \textbf{15,371} pairs of cooperative frames captured at \textbf{10Hz} across \textbf{95} scenes. 
Each frame pair contains synchronized sensor data from both vehicle-side and infrastructure-side perspectives. 
We conduct training on \textbf{7,445} frame pairs from \textbf{46} scenes and evaluate on \textbf{3,316} frame pairs from \textbf{21} validation scenes. 
This dataset is used for comparative experiments and ablation studies, with results reported on the validation set.

\textbf{V2X-Sim}~\cite{li2022v2x} is a simulation dataset collected from virtual environments in \textbf{Carla}. 
It comprises \textbf{10,000} cooperative frames recorded at \textbf{5Hz} across \textbf{100} scenes. 
To validate the generalizability of our approach, we conduct VIC temporal perception experiments on this dataset. 
In each scene, we select one \textbf{road-side unit (RSU)} and one \textbf{CAV} to establish cooperative perception conditions. 
We train on \textbf{5,000} frame pairs from \textbf{50} scenes and evaluate on \textbf{2,000} frame pairs from \textbf{20} validation scenes.

\subsubsection{LET-VIC Implementation Details}\label{sec:let-vic implementation details}
We extract V2X-Seq-SPD~\cite{yu2023v2x} at a frequency of 10Hz to create the V2X-Seq-SPD-10Hz dataset~\cite{yu2023v2x}, aligning with nuScenes~\cite{caesar2020nuscenes}. 
The ego vehicle-side BEV range configuration is \([-51.2, 51.2, -51.2, 51.2]\) meters with grids sized at 0.512 meters $\times$ 0.512 meters. 
Considering the infrastructure-side LiDAR's fan-shaped perception, the infrastructure-side BEV range configuration is \([0, 102.4, -51.2, 51.2]\) meters with grids sized at 0.512 meters $\times$ 0.512 meters. 
We also evaluate our method on the V2X-Sim dataset~\cite{li2022v2x}, which is extracted at 5Hz.  
To establish cooperative perception conditions, we select one RSU and one CAV per scene.  
The ego vehicle-side BEV range configuration remains the same as in V2X-Seq-SPD~\cite{yu2023v2x}, while the infrastructure-side BEV range configuration is set to \([-51.2, 51.2, -51.2, 51.2]\) meters, with the same grid resolution of 0.512m $\times$ 0.512m.
Our evaluation encompasses all tasks utilizing cooperative-view annotations, and the experiments are conducted using the NVIDIA A100 GPU.
To assess the computational feasibility of LET-VIC, we analyze its complexity in terms of floating point operations per second (FLOPs) and parameter count. 
LET-VIC requires 514.86 G FLOPs and consists of 30.11 M parameters. 
These values reflect the model's capability to efficiently leverage vehicle-infrastructure cooperation while maintaining practical computational overhead. 
The balance between accuracy and efficiency makes LET-VIC well-suited for real-world deployment.

\begin{table*}[htpb!]
    \centering
    \small
    \caption{\textbf{Performance Comparison under Different Delay Conditions on the V2X-Seq-SPD-10Hz Dataset} 
    V2VNet~\cite{wang2020v2vnet}, FFNet~\cite{yu2023flow}, and PointPillars~\cite{lang2019pointpillars} follows the tracking-by-detection paradigm with multi-view point cloud feature fusion, 
    LET-V is an end-to-end tracking model with single-view point cloud data input, 
    LET-VIC is an end-to-end tracking model that fuses multi-view point cloud data feature. 
    Detection uses mAP and tracking uses AMOTA and AMOTP.}
    \label{tab:VIC Tracking Benchmark results}
    \renewcommand\arraystretch{1.25}
    \scalebox{0.9}{
    \begin{tabular}{>{\centering\arraybackslash}p{2.5cm}|>{\centering\arraybackslash}p{0.8cm}>{\centering\arraybackslash}p{1.1cm}>{\centering\arraybackslash}p{1.1cm}|>{\centering\arraybackslash}p{0.8cm}>{\centering\arraybackslash}p{1.1cm}>{\centering\arraybackslash}p{1.1cm}|>{\centering\arraybackslash}p{0.8cm}>{\centering\arraybackslash}p{1.1cm}>{\centering\arraybackslash}p{1.1cm}|>{\centering\arraybackslash}p{0.8cm}>{\centering\arraybackslash}p{1.1cm}>{\centering\arraybackslash}p{1.1cm}}
    \hline
    \hline
    \multirow{2}{*}{Model} & \multicolumn{3}{c|}{Latency 0ms} & \multicolumn{3}{c|}{Latency 100ms} & \multicolumn{3}{c|}{Latency 200ms} & \multicolumn{3}{c}{Latency 300ms} \\
    & mAP$\uparrow$ & AMOTA$\uparrow$ & AMOTP$\downarrow$ & mAP$\uparrow$ & AMOTA$\uparrow$ & AMOTP$\downarrow$ & mAP$\uparrow$ & AMOTA$\uparrow$ & AMOTP$\downarrow$ & mAP$\uparrow$ & AMOTA$\uparrow$ & AMOTP$\downarrow$ \\
    \hline
    LET-V (Ours) & 0.456 & 0.467 & 1.113 & - & - & - & - & - & - & - & - & - \\
    \hline
    V2VNet~\cite{wang2020v2vnet} & 0.349 & 0.377 & 1.258 & 0.346 & 0.386 & 1.268 & 0.341 & 0.386 & 1.268 & 0.333 & 0.362 & 1.309 \\
    \hline
    FFNet~\cite{yu2023flow} & 0.349 & 0.377 & 1.258 & 0.340 & 0.377 & 1.261 & 0.345 & 0.383 & 1.262 & 0.342 & 0.387 & 1.264 \\
    \hline
    PointPillars~\cite{lang2019pointpillars} & 0.469 & 0.509 & 0.822 & 0.456 & 0.510 & 0.833 & 0.444 & 0.486 & 0.875 & 0.441 & 0.478 & 0.874 \\
    \hline
    \textbf{LET-VIC (Ours)} & \textbf{0.606} & \textbf{0.640} & \textbf{0.700} & \textbf{0.579} & \textbf{0.631} & \textbf{0.716} & \textbf{0.554} & \textbf{0.613} & \textbf{0.775} & \textbf{0.526} & \textbf{0.570} & \textbf{0.811} \\
    \hline
    \hline
    \end{tabular}
    }
\end{table*}

\begin{figure}[t]
    \centering
    \includegraphics[width=0.48\textwidth]{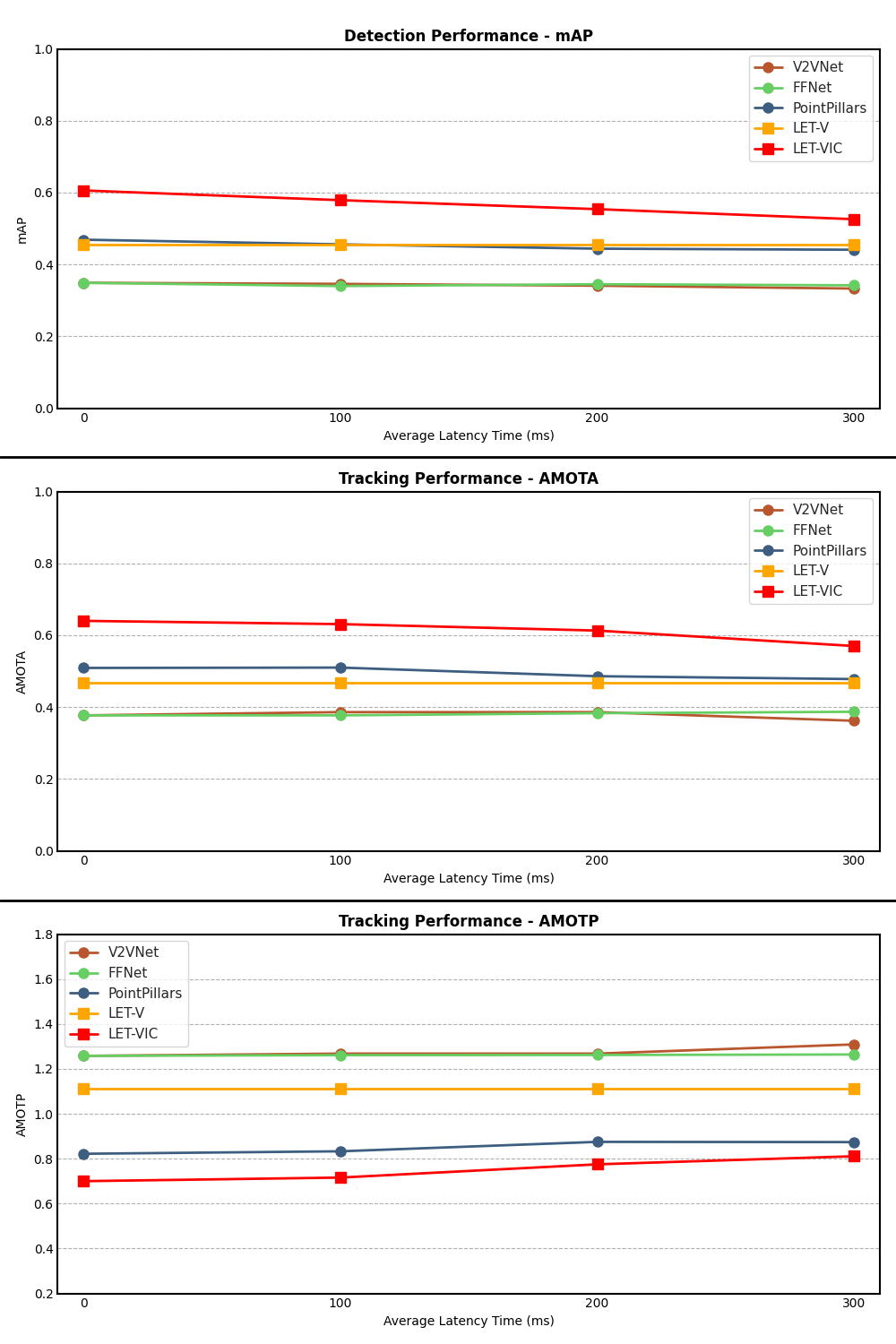}
\vspace{-10pt}
\caption{\textbf{Performance Comparison under Different Latency Conditions.} Our proposed LET-VIC model significantly enhances perception performance through multi-perspective fusion in Vehicle-Infrastructure Cooperation (VIC), achieving the best performance across all metrics and under all latency conditions compared to all baseline models.}
\label{fig: perception_performance_fig}
\end{figure}

\subsubsection{Baseline Settings}\label{sec:vic baseline settings}
We select single-vehicle temporal perception and non-end-to-end VIC perception as two baselines, 
comparing our approach against state-of-the-art (SOTA) LiDAR-based VIC perception methods, including V2VNet~\cite{wang2020v2vnet}, FFNet~\cite{yu2023flow}, and PointPillars~\cite{lang2019pointpillars}.
\begin{itemize}
\item \textbf{LET-V:} Using only ego-vehicle point cloud data as input, LET-V is trained within the LET-VIC framework without incorporating infrastructure data. 
The model is end-to-end and directly outputs detection and tracking results during inference.
\item \textbf{V2VNet~\cite{wang2020v2vnet}:} Detection is performed using V2VNet~\cite{wang2020v2vnet}, paired with AB3DMOT~\cite{weng20203d} for tracking in a tracking-by-detection framework. 
The model is trained on fused features from both ego-vehicle and infrastructure point clouds. Evaluation includes delay scenarios of 0 ms, 100 ms, 200 ms, and 300 ms.
\item \textbf{FFNet~\cite{yu2023flow}:} Detection is performed using FFNet~\cite{yu2023flow}, paired with AB3DMOT~\cite{weng20203d} for tracking in a tracking-by-detection framework. 
The model is trained on fused features from both ego-vehicle and infrastructure point clouds. Evaluation includes delay scenarios of 0 ms, 100 ms, 200 ms, and 300 ms.
\item \textbf{PointPillars~\cite{lang2019pointpillars}:} Detection is performed using PointPillars~\cite{lang2019pointpillars}, paired with AB3DMOT~\cite{weng20203d} for tracking in a tracking-by-detection framework. 
The model is trained on fused features from both ego-vehicle and infrastructure point clouds. Evaluation includes delay scenarios of 0 ms, 100 ms, 200 ms, and 300 ms.
\end{itemize}

\begin{table}[t]
    \centering
    \small
    \setlength{\tabcolsep}{3pt}
    \caption{\textbf{Performance Comparison under Ideal Communication Conditions on the V2X-Seq-SPD-10Hz and V2X-Sim 5Hz Datasets.}}
    \renewcommand\arraystretch{1.25}
    \label{tab:spd and v2x-sim results}
    \scalebox{0.90}{
    \begin{tabular}{c|ccc|ccc}
    \hline
    \hline
    \multirow{2}{*}{Model} & \multicolumn{3}{c|}{V2X-Seq-SPD} & \multicolumn{3}{c}{V2X-Sim} \\
    & mAP$\uparrow$ & AMOTA$\uparrow$ & AMOTP$\downarrow$ & mAP$\uparrow$ & AMOTA$\uparrow$ & AMOTP$\downarrow$ \\
    \hline
    LET-V & 0.456 & 0.467 & 1.113 & 0.187 & 0.093 & 1.734 \\
    \hline
    LET-VIC & \textbf{0.606} & \textbf{0.640} & \textbf{0.700} & \textbf{0.209} & \textbf{0.160} & \textbf{1.606} \\
    \hline
    \hline
    \end{tabular}
    }
\end{table}

\subsection{Experiment Results}\label{sec:experiment results}

\begin{figure*}[ht]
	\centering
	\includegraphics[width=1.0\textwidth]{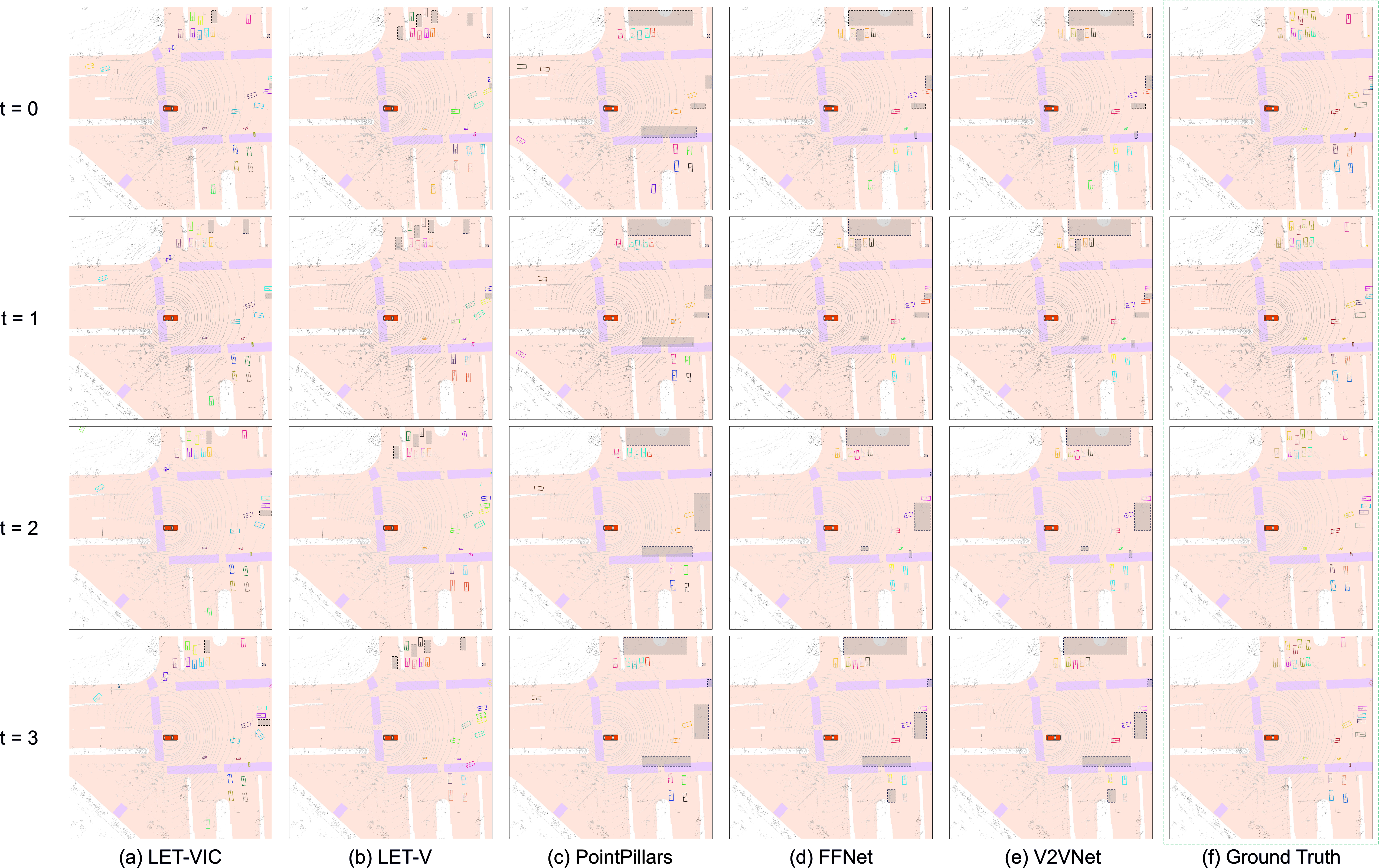}
    \vspace{-10pt}
	\caption{\textbf{Visualization with V2X-Seq-SPD.} 
    The visualization comparison clearly demonstrates that our LET-VIC model achieves superior results compared to other baseline models. Each row represents the perception results of each model at time $t$. 
    Each column shows the temporal perception results of a specific model, with the same color box used for each tracked object. 
    To further illustrate perception performance, we introduce \textbf{semi-transparent gray dashed missed-detection boxes} to highlight undetected objects by each model. LET-VIC exhibits the fewest missed-detection boxes, demonstrating its superior perception capability. 
    Panel (a) shows the temporal perception results of LET-VIC, (b) represents LET-V, (c) shows PointPillars~\cite{lang2019pointpillars}, (d) shows FFNet~\cite{yu2023flow}, (e) shows V2VNet~\cite{wang2020v2vnet}, and (f) represents ground truth.}
	\label{fig:visual-results}
\end{figure*}

Table~\ref{tab:VIC Tracking Benchmark results} and Fig.~\ref{fig: perception_performance_fig} report the comparative detection and tracking performance of different methods on the V2X-Seq-SPD-10Hz Dataset~\cite{yu2023v2x}, where LET-VIC shows superior performance across all metrics. 
The comparison between LET-VIC and LET-V, both in the benchmark results and visualizations, demonstrates the advantage of VIC over single-vehicle perception, underscoring the value of incorporating infrastructure-side information.  
Table~\ref{tab:spd and v2x-sim results} further validates the generalizability of our method through cross-dataset experiments on the V2X-Sim Dataset~\cite{li2022v2x}. 
Moreover, the evaluation and visual results of LET-VIC, along with those of PointPillars~\cite{lang2019pointpillars}, FFNet~\cite{yu2023flow}, and V2VNet~\cite{wang2020v2vnet}, highlight the benefits of end-to-end temporal perception, 
showing that integrating historical temporal information significantly enhances perception performance. 
The inference results of each model are visualized in Fig.~\ref{fig:visual-results}. As shown in the figure, LET-VIC demonstrates robust object tracking across frames, as evidenced by the consistent color-coding of tracked objects, 
while the noticeably fewer semi-transparent gray dashed missed-detection boxes further underscore its enhanced perception capability and reduced detection failures.

\subsubsection{Comparative Results of VIC Perception and Single-Vehicle Perception}
The comparative results of LET-VIC and LET-V demonstrate that LET-VIC consistently outperforms LET-V under various delay conditions, highlighting its strong robustness to latency. 
This improvement is primarily attributed to vehicle-infrastructure coorperation, where infrastructure-side sensors supply complementary perception data, effectively mitigating blind spots and occlusions in the vehicle's field of view. 
Through the VIC Cross-Attention module, LET-VIC seamlessly integrates multi-view BEV features from infrastructure sensors, bridging information gaps in the single-vehicle perspective.
This integration significantly enhances detection accuracy and tracking performance while maintaining robustness, even under delayed conditions.
Specifically, in detection performance, LET-VIC achieves substantial improvements over LET-V, with \textbf{+15.0\%} mAP at 0ms, \textbf{+12.3\%} mAP at 100ms, \textbf{+9.8\%} mAP at 200ms and \textbf{+7.0\%} mAP at 300ms. 
For tracking performance, LET-VIC outperforms LET-V with \textbf{+17.3\%} AMOTA at 0 ms, \textbf{+16.4\%} AMOTA at 100 ms, \textbf{+14.6\%} AMOTA at 200 ms and \textbf{+10.3\%} AMOTA at 300 ms. 
Moreover, LET-VIC consistently achieves lower AMOTP values than LET-V, with \textbf{-0.413} AMOTP at 0 ms, \textbf{-0.397} AMOTP at 100 ms, \textbf{-0.338} AMOTP at 200 ms and \textbf{-0.302} AMOTP at 300 ms, reflecting superior localization accuracy. 
Additionally, as shown in Fig.~\ref{fig: perception_performance_fig}, although each metric shows a slight decline within a latency range up to 300 ms, LET-VIC still surpasses LET-V by
\textbf{+7.0\%} in mAP, \textbf{+10.3\%} in AMOTA, and \textbf{-0.302} in AMOTP at 300 ms delay, 
demonstrating its ability to effectively fuse infrastructure features under certain delays and enhance perception capabilities, showcasing robustness to latency.
Furthermore, our experiments on the V2X-Sim dataset~\cite{li2022v2x} validate the generalizability of our approach under more challenging cooperative conditions. 
Unlike the V2X-Seq-SPD dataset~\cite{yu2023v2x}, which offers dense collaboration with a 10Hz sampling frequency and full-frame cooperation, 
V2X-Sim~\cite{li2022v2x} operates at a lower temporal resolution (5Hz) and exhibits sparser cooperation—only 87.2\% of frames contain valid collaborative perception due to limited sensor overlap between the RSU and CAV. 
Despite these constraints, LET-VIC still achieves notable improvements over LET-V, with an increase of \textbf{+2.2\%} in mAP and \textbf{+6.7\%} in AMOTA. 
This performance gain under partial cooperation underscores LET-VIC's capability to effectively leverage intermittent infrastructure signals and maintain robust perception even when continuous collaboration is not feasible. 
Overall, the consistent performance across these diverse datasets confirms the adaptability of our approach to different sensor configurations and collaboration frequencies.

\subsubsection{Comparative Results of End-to-End Paradigms and Tracking-by-Detection Paradigms}
The comparative results between LET-VIC and models such as V2VNet~\cite{wang2020v2vnet}, FFNet~\cite{yu2023flow}, and PointPillars~\cite{lang2019pointpillars} indicate that the end-to-end temporal perception approach of LET-VIC significantly outperforms traditional tracking-by-detection models across various delay settings. 
The performance improvement of LET-VIC is due to its ability to fuse temporal history information. 
This allows previously visible objects' features to be carried forward, compensating for occlusions in the current frame. For example, targets visible in prior frames but obscured in the current frame benefit from temporal feature integration, resulting in enhanced detection and tracking performance. 
This capability is particularly beneficial in addressing blind spots in the current frame by leveraging features from previous observations.
Specifically, LET-VIC achieves substantial improvements in detection performance, with at least \textbf{+13.7\%} mAP at 0ms, \textbf{+12.3\%} mAP at 100ms, \textbf{+11.0\%} mAP at 200ms and \textbf{+8.5\%} mAP at 300ms compared to the other models. 
For tracking performance, LET-VIC outperforms V2VNet~\cite{wang2020v2vnet}, FFNet~\cite{yu2023flow}, and PointPillars~\cite{lang2019pointpillars} with at least \textbf{+13.1\%} AMOTA at 0 ms, \textbf{+12.1\%} AMOTA at 100 ms, \textbf{+12.7\%} AMOTA at 200 ms and \textbf{+9.2\%} AMOTA at 300 ms. 
Moreover, LET-VIC consistently achieves lower AMOTP values, with at least \textbf{-0.122} AMOTP at 0 ms, \textbf{-0.117} AMOTP at 100 ms, \textbf{-0.100} AMOTP at 200 ms and \textbf{-0.063} AMOTP at 300 ms, reflecting superior localization accuracy. 
Additionally, Compared to V2VNet~\cite{wang2020v2vnet}, FFNet~\cite{yu2023flow}, and PointPillars~\cite{lang2019pointpillars}, LET-V demonstrates that even with single-vehicle perception, the end-to-end approach significantly improves detection and tracking.
Its performance approaches that of PointPillars~\cite{lang2019pointpillars} and surpasses both FFNet~\cite{yu2023flow} and V2VNet~\cite{wang2020v2vnet}. This highlights the effectiveness of the end-to-end paradigm in enhancing perception performance, even without infrastructure data.

\subsection{Ablation Study}\label{sec:ablation study}

\begin{table}[t]
    \centering
    \small
    \caption{\textbf{Importance of CEC in LET-VIC on the V2X-Seq-SPD-2Hz Dataset at Different Latency Levels.}}
    \renewcommand\arraystretch{1.25}
    \label{tab:SC Tracking Benchmark results}
    \scalebox{0.90}{
    \begin{tabular}{c|c|ccc}
    \hline
    \hline
    \textbf{CEC} & \textbf{Latency (ms)} & \textbf{mAP~$\uparrow$} & \textbf{AMOTA~$\uparrow$} & \textbf{AMOTP~$\downarrow$} \\
    \hline
    & \multirow{2}{*}{0}& 0.198 & 0.092 & 1.653 \\
    \checkmark & & \textbf{0.262} & \textbf{0.139} & \textbf{1.615} \\
    \hline
    & \multirow{2}{*}{100} & 0.206 & 0.120 & 1.647 \\
    \checkmark & & \textbf{0.281} & \textbf{0.175} & \textbf{1.568} \\
    \hline
    & \multirow{2}{*}{200} & 0.204 & 0.118 & 1.652 \\
    \checkmark & & \textbf{0.254} & \textbf{0.136} & \textbf{1.608} \\
    \hline
    & \multirow{2}{*}{300} & 0.191 & 0.099 & 1.653 \\
    \checkmark & & \textbf{0.243} & \textbf{0.141} & \textbf{1.611} \\
    \hline
    \hline
    \end{tabular}
    }
\end{table}

This section investigates the impact of our CEC module on perception accuracy. Calibration errors can significantly degrade tracking performance, and our module is designed to mitigate these effects.

\subsubsection{Experimental Setup}\label{sec:sc experimental setup}

We compare the performance of the LET-VIC model with and without the CEC module under delays of 0 ms, 100 ms, 200 ms, and 300 ms. The dataset annotations are used without manual error compensation to highlight the module's effectiveness.

\subsubsection{LET-VIC Implementation Details}\label{sec:sc LET-VIC implementation details}
Due to limited computational resources, this section of experiments extracts the V2X-Seq-SPD dataset~\cite{yu2023v2x} at a reduced frequency of 2Hz to generate the V2X-Seq-SPD-2Hz dataset~\cite{yu2023v2x}. 
Additionally, we reduced the feature dimensionality in the model from 256 dimensions to 64 dimensions to expedite training and evaluation. 
Other settings remain consistent with those described in Section~\ref{sec:let-vic implementation details}.

\subsubsection{Analysis}\label{sec:ablation study analysis}
The results presented in Table~\ref{tab:SC Tracking Benchmark results} demonstrate that LET-VIC with the CEC module consistently outperforms the standard LET-VIC across all delay settings, highlighting the CEC module's effectiveness in enhancing perception robustness.
By compensating for sensor calibration errors, the CEC module improves the accuracy of multi-view feature fusion. This ensures that the LiDAR feature points from the vehicle and infrastructure sides, corresponding to the same real-world location, are correctly aligned. 
Without such compensation, misalignments between vehicle-side and infrastructure-side feature points may introduce noise during feature fusion, negatively impacting perception performance.
Specifically, LET-VIC with calibration error compensation achieves \textbf{+6.4\%} mAP and \textbf{+4.7\%} AMOTA improvements, along with a \textbf{-0.038} AMOTP reduction at 0 ms delay; 
\textbf{+7.5\%} mAP, \textbf{+5.5\%} AMOTA, and \textbf{-0.079} AMOTP at 100 ms delay; 
\textbf{+5.0\%} mAP, \textbf{+1.8\%} AMOTA, and \textbf{-0.044} AMOTP at 200 ms delay; 
and \textbf{+5.2\%} mAP, \textbf{+4.2\%} AMOTA, and \textbf{-0.042} AMOTP at 300 ms delay.
By compensating for sensor calibration errors, this module improves multi-view feature fusion, leading to better overall performance and demonstrating LET-VIC's robustness to calibration inaccuracies.

\section{Conclusion}
This paper introduces LET-VIC, a LiDAR-based end-to-end tracking framework for VIC. 
By integrating information from multi-view sensor sources, LET-VIC addresses the limitations of single-view perception and enhances the system's ability to detect and track objects. 
Through its end-to-end design, LET-VIC effectively incorporates temporal information to maintain object identity across frames in both detection and tracking, significantly improving performance in dynamic environments.
Moreover, its feature fusion approach compensates for spatial calibration errors during vehicle-infrastructure coordinate transformations, ensuring robust and accurate object perception. 
Experimental results on the V2X-Seq-SPD dataset confirm the effectiveness of this approach, demonstrating substantial improvements in tracking performance and system robustness.

LET-VIC, while effective, is limited by relying solely on point cloud inputs from LiDAR sensors. 
Although point clouds provide accurate depth and 3D structure, they lack color and texture details, affecting the recognition of fine object attributes. 
Additionally, point clouds can become sparse at longer distances or in complex environments, impacting detection and tracking accuracy. 
Conversely, image data offers rich semantic information but lacks depth. 
Future work will focus on fusing point cloud and image features to enhance both spatial perception and semantic understanding, improving overall system performance in complex environments.
Another limitation is the lack of compensation for communication delays in VIC, which can affect real-time tracking performance. 
Future research will address this by developing strategies to mitigate temporal delays, ensuring more accurate and robust tracking across multi-view sensors.

% REFERENCES
{
\small
\bibliographystyle{IEEEtran}
\bibliography{IEEEabrv,reference}
}

\end{document}